\documentclass[10pt,twocolumn,letterpaper,hidelinks]{article}

\usepackage{cvpr}
\usepackage{times}
\usepackage{epsfig}
\usepackage{graphicx}
\usepackage{amsmath}
\usepackage{amssymb}
\usepackage{algorithm}
\usepackage{algorithmic}
\usepackage{tabularx}
\usepackage{soul}
\usepackage{multirow}

\usepackage[breaklinks=true,bookmarks=false]{hyperref}

\newcommand{\ttinneriter}{\texttt{inner\_iters}}
\newcommand{\ttit}{\texttt{iter}}
\newcommand{\ttmaxiter}{\texttt{max\_iter}}
\newcommand{\ttbestScore}{\texttt{best\_score}}

\newcommand{\cM}{\mathcal{M}}

\newcommand{\cS}{\mathcal{S}}
\newcommand{\cD}{\mathcal{D}}
\newcommand{\cP}{\mathcal{P}}
\newcommand{\cQ}{\mathcal{Q}}

\newcommand{\bX}{\mathbf{X}}
\newcommand{\bY}{\mathbf{Y}}
\newcommand{\bZero}{\mathbf{0}}

\newcommand{\hbX}{\hat{\mathbf{X}}}

\newcommand{\bs}{\mathbf{s}}

\newcommand{\bp}{\mathbf{p}}
\newcommand{\bq}{\mathbf{q}}

\newcommand{\bd}{\mathbf{d}}

\newcommand{\bA}{\mathbf{A}}

\newcommand{\bR}{\mathbf{R}}
\newcommand{\bt}{\mathbf{t}}

\newcommand{\bbR}{\mathbb{R}}
\newcommand{\cC}{\mathcal{C}}
\newcommand{\Nsample}{$N_\text{sample}$}

\cvprfinalcopy 

\def\cvprPaperID{494} 

\begin{document}

\title{SDRSAC: Semidefinite-Based Randomized Approach \\ for Robust Point Cloud Registration without Correspondences }

\author{Huu M. Le$^{1}$, Thanh-Toan Do$^{2,3}$, Tuan Hoang$^{1}$, and Ngai-Man Cheung$^{1}$\\
{\small $^{1}$Singapore University of Technology and Design} 
{\small $^{2}$University of Liverpool}
{\small $^{3}$AIOZ Pte Ltd}
\\
}

\maketitle

\begin{abstract}
    This paper presents a novel randomized algorithm for robust point cloud registration without correspondences. Most existing registration approaches require a set of putative correspondences obtained by extracting invariant descriptors. However, such descriptors could become unreliable in noisy and contaminated settings. In these settings, methods that directly handle input point sets are preferable.  Without correspondences, however,   conventional randomized techniques require a very large amount of samples in order to reach satisfactory solutions. In this paper, we propose a novel approach to address this problem. In particular, our work enables the use of randomized methods for point cloud registration without the need of putative correspondences. By considering point cloud alignment as a special instance of graph matching and employing an efficient semi-definite relaxation, we propose a novel sampling mechanism, in which the size of the sampled subsets can be larger-than-minimal. Our tight relaxation scheme enables fast rejection of the outliers in the sampled sets, resulting in high quality hypotheses. We conduct extensive experiments to demonstrate that our approach outperforms other state-of-the-art methods. Importantly, our proposed method serves as a generic framework which can be extended to problems with known correspondences~\footnote{Source code is available at: https://github.com/intellhave/SDRSAC.}.
\end{abstract}


\section{Introduction}

Point cloud registration is an important problem in many computer vision applications, including range scan alignment~\cite{lu1997globally}, 3D object recognition and localization~\cite{drost2010model,tran2019device}, large scale reconstruction~\cite{agarwal2011building, schonberger2016structure}. Given two sets of points in three-dimensional (3D) Euclidean space, the objective is to search for the optimal rigid transformation, which comprises a rotation matrix $\bR^* \in SO(3)$ and a translation vector $\bt^* \in \bbR^3$, that optimally aligns the two input point sets. In many practical applications, the input data contains a significant amount of noise. Moreover, the overlapping region between the two point sets can be small, resulting in a large number of outliers, i.e., non-overlapping points. Therefore, the registration needs to be conducted in a robust manner so that the final estimates are not affected by the contamination. Formally, let $\cS = \{\bs_i \in \bbR^3 \}_{i=1}^{N_s}$ and $\cD = \{ \bd_j \in \bbR^3 \}_{j=1}^{N_d}$ denote the source and destination (target) point clouds, respectively, the problem of robust rigid registration can be formulated as

\begin{equation} 
	\label{eq_prob_def}
	\min_{\bR \in SO(3), \bt \in \bbR^3} \sum_{i=1}^{N_s} \rho\left( \|\bR \bs_i + \bt - \bd_j\|\right),
\end{equation}
where the notation $\|\cdot\|$ represents the $\ell_2$ norm,  $\bd_j$ is a point in the target set $\cD$ that is closest to the transformed point $\bR\bs_i + \bt$, i.e., 
\begin{equation}
\label{eq_nearest_point}
 \bd_j = \arg\min_{\bd_k \in \cD} \|\bR \bs_i + \bt - \bd_k\|,
\end{equation}
and $\rho$ is a robust loss function. Here, $SO(3)$ denotes the space of rotation matrices. In order for the registration to be robust, $\rho$ is typically chosen from a set of robust kernels~\cite{huber81,aftab2015convergence,lenon}. In this work, we make use of the popular maximum consensus criterion~\cite{hartley2003multiple}, in which $\rho$ is defined as
\begin{equation}
\label{eq_rho}
\rho (x) = \begin{cases}
0 \;\; \text{if} \;\;\;  x \le \epsilon, \\
1 \;\; \text{otherwise}.
\end{cases}	
\end{equation}
The threshold $\epsilon > 0$ is a user-defined parameter that specifies the maximum allowable distance for a correspondence pair to be considered as an inlier. Intuitively, by solving~\eqref{eq_prob_def} with $\rho$ defined as per~\eqref{eq_rho}, we search for the optimal alignment $(\bR^*, \bt^*)$ that maximizes the set of overlapping points, where $\bs_i \in \cS$ overlaps $\bd_j \in \cD$ if the transformation $(\bR^*, \bt^*)$  brings $\bs_i$ to a new location that lies within the ball $(\bd_j, \epsilon)$. The problem~\eqref{eq_prob_def} is an active research topic in computer vision due to its computational complexity. 

Many existing algorithms need to take as input a set of putative correspondences, which are usually obtained (as a pre-processing step) by extracting local invariant features on the given point sets, and executing multiple routines of nearest neighbor search to propose initial key-point matches~\cite{aldoma2012tutorial, gal2006salient}. Several types of 3D local features~\cite{scovanner20073,chua1997point,zhang2014simplex,rusu2009fast}  have demonstrated to provide promising results throughout a variety of challenging datasets. However, noise and contamination would degrade the quality of extracted features. Furthermore, in order for the local features to be precisely computed, many feature extractors require the surface representation to be dense; however, sparse point clouds are common in practice. In particular, it has been demonstrated in~\cite{aiger20084} that, for noisy datasets with high proportion of outliers,  alignments using features have poorer results compared with using raw data. Therefore, there is interest to develop registration algorithms that directly align raw point cloud data without the need of a priori correspondences~\cite{bustos2016fast, aiger20084, breuel2003implementation,li20073d, enqvist2009optimal}. 

Due to the computational complexity of the problem under different input settings, randomized hypothesize-and-verify algorithms such as RANSAC~\cite{fischler1981random} and its variants~\cite{chum2003locally,chum2005matching,torr2000mlesac,tordoff2005guided,raguram2013usac,le2017ratsac} are popular approaches. Randomized techniques have been employed to address problems with known correspondences. On the other hand, similar random sampling strategy as that of RANSAC can also be applied to problems of  without correspondences. Specifically, at each iteration, a minimal subset (of three points) on each point cloud can be sampled to form three pairs of correspondences, which are used for estimating and validating one hypothesis. Such procedure can then be repeated until a satisfactory solution is obtained. However, with noise and outliers, the likelihood of picking outlier-free subsets degrades rapidly. Therefore, much efforts have been devoted to develop better sampling strategies, notably the 4-Points Congruent Sets (4PCS) method proposed in~\cite{aiger20084} and its improved variant~\cite{mellado2014super}. Although 4PCS provides considerable advantages over  conventional randomized methods, the enumeration of (approximately) congruent sets that underpins this algorithm is the main issue when working with point clouds with large number of points and high outlier rates. In fact, for dense point clouds, 4PCS and its variants need to down-sample the point cloud before conducting the sampling process to reduce processing time.

In this paper, we address the above-mentioned limitations. Specifically, by employing a special instance of graph matching formulation, we propose a new \emph{larger-than-minimal sampling} strategy for point cloud registration without correspondences. The advantage of our method is that the task of searching for correspondences is quickly approximated by solving a relaxed convex problem, rather than subset enumeration.  This allows us to sample subsets with arbitrarily large size, in which sets of correspondences are obtained from solutions of convex semi-definite programming. These correspondences can then be used for estimating and validating hypotheses. A large subset of points in the source point cloud represent better its structure, and by identifying its corresponding points on the target set, the two point clouds can be coarsely align faster, which can then be refined by local methods, e.g., ICP~\cite{besl1992method}. Empirical results show that the proposed method is very competitive. {\bf Our main contributions are:} 
\begin{itemize}
	\item We apply graph matching for the registration problem without correspondences using a novel cost function to enforce robustness and a tight semidefinite (SDP) relaxation.
	\item From the SDP formulation, we then develop a new \emph{larger-than-minimal} subset sampling scheme, leading to an effective randomized algorithm that outperforms other state-of-the-art methods.	
\end{itemize}
\section{Related Work}
\label{sec_related_work}
In practice, if the two input point clouds are coarsely aligned, the well-known Iterative Closest Point (ICP)~\cite{besl1992method} is commonly employed. Like other iterative algorithms, ICP alternates between establishing correspondences and estimating the transformation. 
The main drawback of this method lies in the fact that it requires a good starting point (initial pose) to prevent itself from converging to poor alignments. Additionally, in terms of robustness, ICP suffers from the same weakness as that of the least squares estimator, i.e., it is easily biased by erroneous outliers. Several works~\cite{chetverikov2002trimmed,bergstrom2017robust,phillips2007outlier} have been proposed to improve the such shortcomings of ICP. However, these variants still need to be bootstrapped by a decent initial pose. ICP is therefore commonly employed as a local refinement procedure, which is executed after the point clouds are roughly registered by some type of global alignment algorithms.

Algorithms that offer globally optimal solutions are also actively developed in the literature. To address the initialization-dependent issue of ICP,  Go-ICP~\cite{yang2016go} employs the branch and bound strategy to search in the space of $\bR$ and $\bt$ for the optimal transformation. However, due to the least squares objective of Go-ICP, its returned optimal solutions are still non-robust to outliers. The robustness of Go-ICP can be improved by incorporating robust loss functions in place of least squares. Another globally optimal algorithm to tackle the robust registration without correspondences was proposed by Bustos et al.~\cite{bustos2016fast}. Unlike Go-ICP,~\cite{bustos2016fast} solves~\eqref{eq_prob_def} directly and its solution is robust to outliers. Branch and bound is also the mechanism behind~\cite{bustos2016fast}, with a novel steographic projection implementation for fast matching query and tighter bound computation. Although the convergence to globally optimal solutions is guaranteed for these methods, they are still impractical for large datasets due to their expensive computational cost.

As previously mentioned, in some point cloud registration applications, randomized strategies -- with the famous RANSAC~\cite{fischler1981random} representative -- are still the dominant approaches. Our work also belongs to this category. Generally speaking, for most randomized methods, the underlying sampling strategy significantly affects the run time, since the termination of a sampling mechanism depends largely on its ability to quickly identify outlier-free subsets. For the case with no correspondences, it is even harder since two outlier-free subsets, one in each input point cloud, need to be identified \emph{and} the elements in these two subsets must form correct correspondences. Different sampling strategies have been proposed in the literature~\cite{chum2003locally,chum2005matching,tran2014sampling}. For instance, in problems with known correspondences, one of the notable improvements for RANSAC is LO-RANSAC (Locally Optimal RANSAC)~\cite{chum2003locally}, which proposes to sample \emph{larger-than-minimal} subsets when RANSAC solution is updated. This improved strategy has been shown to significantly outperform conventional RANSAC. Randomized methods can also be improved using local refinement techniques~\cite{le2017exact,cai2018deterministic,purkait2017maximum}.

In the context of robust point cloud alignment without correspondences, however, the idea of larger-than-minimal sampling has yet to be thoroughly explored. Unlike the case of known correspondences, the ``inlier" set at each RANSAC iteration may not be true inliers, and purely applying LO-RANSAC to these subsets may not be of any help, while the run time is increased. Besides, even if true inliers reside in the subset, a larger-than-minimal subset may be contaminated with outliers, which can deteriorate the estimation since it is done solely by solving least squares over the sampled subsets. From the insight discussed above, it can be seen that from any larger-than-minimal subsets, if the outliers are efficiently rejected, the ability to discover good hypotheses can be accelerated. This idea is analyzed in our paper, in which we propose a new algorithm that enables the sampling of any arbitrarily large subsets. Such sampling scheme is incorporated with an oracle that allows the outliers in the sampled point sets to be efficiently discarded. Our experiments show that this newly-proposed algorithm outperforms previous approaches. 

Our work is closely related to 4PCS~\cite{agarwal08}, which is the state-of-the-art sampling approach that also solves the same problem as ours.  Instead of randomly picking minimal subsets of three points, 4PCS works on sampled tuples of four co-planar points. This method is later improved by Super4PCS~\cite{mellado2014super}, where the complexity of congruent set extraction is reduced from quadratic to linear with respect to the number of points. Though efficient, especially with the improvements introduced in~\cite{mellado2014super}, 4PCS still suffers from the same weakness as RANSAC, e.g., the low likelihood of quickly sorting out the right subsets under the presence of high outliers. In fact, for each subset of 4 co-planar points on one point cloud, 4PCS needs to enumerate all congruent subsets on the other point cloud. Therefore, this method needs to subsample the point set into a smaller subset before the sampling is conducted. 

The technique developed in this work is also inspired by the class of methods that solve 3D registration problem using graph matching~\cite{enqvist2009optimal,leordeanu2005spectral,le2017alternating} and recent semi-definite relaxation for non-rigid registration~\cite{dym2017dspp,hullinefficient,kezurer2015tight}. These algorithms, however, can only work for a small number of data points, which is very inefficient if applied directly on large point clouds. We, on the other hand, propose to employ graph matching as sub-problems embedded in a random sampling framework. This allows us to combine the strengths of both classes of methods, i.e., randomized techniques and graph matching, to derive an efficient sampling mechanism for point cloud registration.  

\section{Semidefinite-Based Randomized Approach}
\subsection{The Correspondence Problem}
\label{sec_SDP}
When the putative correspondences are not given, the task of solving~\eqref{eq_prob_def} can be viewed as jointly estimating the transformation $(\bR^*, \bt^*)$ and the best subset of correspondences $\cC^*$. Hence, if $\cC^*$ is provided, the optimal alignment can be computed in a straightforward manner, and vice versa. With that in mind, in this section, we first introduce the correspondence problem to search for $\cC^*$. In the latter sections, we then discuss its semidefinite (SDP) relaxation, which can be employed as a fast hypothesis generation mechanism that lies at the core of our efficient randomized algorithm.

To simplify the formulation, let us for now assume that we are given two 3D point sets $\cP=\{\bp_i\}_{i=1}^N$ and $\cQ = \{\bq_j \}_{j=1}^N$, each contains $N$ data points. The task is to find the optimal set of correspondences $\cC_{\cP\cQ}^*$  that can be used to robustly align $\cP$ and $\cQ$. Note that we use different notations for input data here compared to~\eqref{eq_prob_def}, since this new problem will be used as sub-problem for solving~\eqref{eq_prob_def}, which will be clarified in the latter sections.

Let $\bX \in \{0,1\}^{N \times N}$ be the permutation matrix in which the element at the $i-$th row and $j-$th column (denoted by $\bX_{i,j} $) is assigned the value of $1$ if the pair $\bp_i \in \cP$ and $\bq_j \in \cQ$ belongs to the set of correspondences and $0$ otherwise. To account for outliers, let us further assume that the optimal solution contains  $m \le N$ pairs of correspondences. The value of $m$ can be chosen to be greater than or equal to $3$ (the size of the minimal subset), or estimated based on the known outlier ratio of the problem at hand. Note that with the introduction of $m$, $\bX$ is now a sub-permutation matrix. Denote by $\hat{\bX}$ the vectorization of the matrix $\bX$ obtained by stacking its columns:
\begin{equation}
\hat{\bX} = [\bX_{:,1}^T \;\; \bX_{:,2}^T \;\;\dots \;\; \bX_{:,N}^T ]^T,
\end{equation}
where $\bX_{:,j}$ denotes the $j-$th column of $\bX$. 
In order to search for the best correspondence assignments, consider the following optimization problem:
\begin{subequations}
\label{eq_permuataion}
\begin{align}
& \max_{\bX}  && \hbX^T \bA \hbX, \\
& \text{subject to} && \bX_{i,j} \in \{0,1\} \;\; \forall i, j \in \{1, \dots, N\}, \label{eq_binary_constraint}\\
& && \sum_{j=1}^N \bX_{i,j} \le 1 \;\; \forall i \in \{1, \dots, N\}, \label{eq_sumrow_constraint}\\
& && \sum_{i=1}^N \bX_{i,j} \le 1 \;\; \forall j \in \{1, \dots, N\}, \label{eq_sumcolumn_constraint}\\
& && \sum_{i=1}^N\sum_{j=1}^N \bX_{i,j} = m, \;\; \label{eq_sumxij_constraint}
\end{align}
\end{subequations}
where $\bA \in \bbR^{N^2 \times N^2}$ is the symmetric matrix that characterizes the matching potentials between pairs of points (line segments) in $\cP$ and $\cQ$. In particular, we define the elements of the matrix $\bA$ as

	\begin{equation}
	\label{eq_def_weight_matrix}
	\bA_{ab,cd} = \begin{cases}
	\begin{aligned}
		
	& f(\bp_a, \bp_c, \bq_b, \bq_d) \;\text{if}  |\delta(\bp_a, \bp_c) - \delta(\bq_b, \bq_d)| \le \gamma, &&\\
	&0  \;\;\;\;\;\;\;\;\;\;\;\;\;\;\;\;\;\;\;\;\;\;\;\;\text{otherwise},&&
	\end{aligned}
	\end{cases}
	\end{equation}
where $a, b, c, d\in \{1..N\}$ are point indexes, $ab = a + (b-1)N$ and $cd = c + (d-1)N$ are the indexes for the row and column of $\bA$, respectively;  $\bp_a, \bp_c \in \cP$; $\bq_b, \bq_d \in \cQ$; $\gamma > 0$ is a predefined threshold and $\delta (\bp_1,\bp_2)$ computes the Euclidean distance between two 3D points $\bp_1$ and $\bp_2$; $f$ is a function that takes as input two pairs of points $(\bp_a,\bp_c)$ and $(\bq_b, \bq_d)$ and outputs a scalar that represents matching potential for these two pairs. Typically, $f$ in~\eqref{eq_def_weight_matrix} is chosen to be the function that penalizes the difference in length between the two line segments. For simplicity, we choose $f$ to be $\exp(-|\delta(\bp_a, \bp_c) - \delta(\bq_b, \bq_d)|)$.

The constraints in~\eqref{eq_permuataion} are to assure that $\bX$ must lie in the space of permutation matrices. Specifically, besides the binary constraint~\eqref{eq_binary_constraint} to restrict $\bX_{i,j} \in \{0,1\}$, it is also required that each point $\bp_i \in \cP$ can only be assigned to at most one point in $\cQ$ and vice versa, which is reflected by the constraints~\eqref{eq_sumrow_constraint} and~\eqref{eq_sumcolumn_constraint}. Finally, by enforcing~\eqref{eq_sumxij_constraint}, the optimal solution of~\eqref{eq_permuataion} will only contain $m$ pairs of correspondences. The solution to~\eqref{eq_permuataion} provides the optimal assignment such that the sum of matching potentials gained yield from the corresponding pairs is maximized. 

While the use of graph matching has been explored in several rigid and non-rigid registration problems~\cite{enqvist2009optimal, leordeanu2005spectral, kezurer2015tight,hullinefficient}, most previous work consider solving graph matching for the whole input data, which is infeasible for large datasets. This work proposes to solve graph matching on very small subsets of points, then embed them into a random sampling framework.
Additionally, we also propose a better formulation of the matrix $\bA$ for the special case of robust rigid registration. Particularly, the matrix $\bA$ is designed based on the fact that if $\bp_a \in \cP$ corresponds to $\bq_b \in \cQ$ and $\bp_c \in \cP$ corresponds to $\bq_d \in \cQ$, since the transformation is rigid, the lengths of the two line segments $\bp_a\bp_c$ and $\bq_b\bq_d$ must be approximately the same (due to the effect of noise). With the new formulation ~\eqref{eq_def_weight_matrix}, instead of assigning the matching potentials to all the elements of $\bA$, we only allow pairs of segments whose lengths differ by a small value $\gamma$ to be considered as candidate for matching, while pairs having large gap are rejected.


\subsection{Semidefinite Relaxation}
Let us first consider the equivalent formulation of~\eqref{eq_permuataion}.  Let $\bY = \hbX \hbX^T \in \bbR^{N^2\times N^2}$, the problem~\eqref{eq_permuataion} becomes
\begin{subequations}
\label{eq_max_traceWY}
\begin{align}
\centering
&\max_{\bX, \bY} && \text{trace}(\bA\bY), \\
& \text{subject to} && \bY = \hbX \hbX^T, \label{eq_rank1_constraint}\\
& && \text{trace} (\bY) = m, \label{eq_traceY}  \\
& && 0 \le \bX_{i,j} \le 1 \;\; \forall i,j, \label{eq_x_bound} \\
& && \eqref{eq_sumrow_constraint}, \eqref{eq_sumcolumn_constraint}, 
\eqref{eq_sumxij_constraint}.
\end{align}
\end{subequations}
Note that in~\eqref{eq_max_traceWY}, the constraint~\eqref{eq_binary_constraint} can be removed without affecting the equivalence between~\eqref{eq_max_traceWY} and~\eqref{eq_permuataion}. Indeed, since $\text{trace}(\bY) = m$, it must hold that $\sum_{i,j}\bX^2_{i,j} = m$. On the other hand, due to the constraint~\eqref{eq_sumxij_constraint}, $\sum_{i,j}\bX_{i,j} = m$. These conditions result in $\sum_{i,j}\bX^2_{i,j} = \sum_{i,j}\bX_{i,j}$. Also, due to the condition  $0 \le \bX_{i,j} \le 1$, $\bX_{i,j}$ can only be either $0$ or $1$. In other word, $\bX$ can now be constrained in the convex hull of the sub-permutation matrices~\cite{mendelsohn1958convex}. 

By introducing $\bY$, the problem is lifted into the domain of $\bbR^{N^2\times N^2}$ and the binary constraint of $\bX$ can be relaxed without changing the solution of the problem. However, the problem~\eqref{eq_max_traceWY} is still non-convex due to the rank-one constraint~\eqref{eq_rank1_constraint}. In order to approximate~\eqref{eq_max_traceWY}, we employ the common convex relaxation approach, in which~\eqref{eq_rank1_constraint} is relaxed to the semidefinite constraint $\bY - \hbX \hbX^T \succeq \bZero$. Then, we arrive at the following convex optimization problem:
\begin{subequations}
	\label{eq_max_traceWY_relaxed}
	\begin{align}
	& \max_{\bX, \bY} && \text{trace}(\bA\bY), \\
	& \text{subject to} && \bY - \hbX \hbX^T \succeq \bZero, \label{eq_rank1_constraint_relaxed}\\
	& && \eqref{eq_sumrow_constraint}, \eqref{eq_sumcolumn_constraint}, 
	\eqref{eq_sumxij_constraint},\eqref{eq_traceY}, \eqref{eq_x_bound}. \label{eq_permu_convex_hull}
	\end{align}
\end{subequations}
The problem~\eqref{eq_max_traceWY_relaxed} introduced above is a convex semidefinite program (SDP), whose globally optimal solution can be obtained using many off-the-shelf SDP solvers. In this work, we use SDPNAL+~\cite{yang2015sdpnal} throughout all the experiments. 

\subsubsection{Tightening the Relaxation}
As~\eqref{eq_max_traceWY_relaxed} is solved as an approximation for~\eqref{eq_max_traceWY} with rank-one constraint, one would expect that the two solutions to be as close as possible. Inspired by~\cite{kezurer2015tight}, we add the following constraints to~\eqref{eq_max_traceWY_relaxed} to tighten the relaxation:
\begin{equation}
	\label{eq_tighten_constraint}
	\bY_{ab,cd} \le \begin{cases}
		\begin{aligned}
			&0, && \text{if} \;\; a = c, b \neq d,\\
			&0, && \text{if} \;\; b = d, a \neq c,\\
			& \min (\bX_{ab}, \bX_{cd}), && \text{otherwise}.
		\end{aligned}		
	\end{cases}
\end{equation}
The intuition behind~\eqref{eq_tighten_constraint} is that one point in $\cS$ is not allowed to match with more than one point in $\cD$ and vice versa.  Also, since $\bY_{ab,cd} = \bX_{ab}\bX_{cd}$ and these are binary numbers,  it must hold that $\bY_{ab,cd} \le \min(\bX_{ab},\bX_{cd})$. 

In addition, thanks to the special case of robust registration,~\eqref{eq_max_traceWY_relaxed} can be further tightened. Observe that, based on the discussion of formulating the matrix $\bA$, the following constraints can be enforced:
\begin{equation}
\label{eq_tighten_constraint_2}
\bY_{ab,cd}  = 0 \;\;\text{if} \;\; \bA_{ab,cd}  = 0,
\end{equation}
which means if the pairs $\bp_a\bp_c$ and $\bp_b\bp_d$ differ too much in length (more than $\gamma$ , we directly disallow them to be matched. 

Finally, with the addition of~\eqref{eq_tighten_constraint} and~\eqref{eq_tighten_constraint_2}, our SDP relaxation becomes
\begin{subequations}
	\label{eq_max_traceWY_relaxed_tight}
	\begin{align}
	& \max_{\bX, \bY} && \text{trace}(\bA\bY), \\
	& \text{subject to} && \eqref{eq_rank1_constraint_relaxed}, \eqref{eq_traceY},\eqref{eq_x_bound}, \eqref{eq_permu_convex_hull}, \eqref{eq_tighten_constraint}, \eqref{eq_tighten_constraint_2}.
	\end{align}
\end{subequations}
Note that~\eqref{eq_max_traceWY_relaxed_tight} is still a convex SDP since the additional constraints~\eqref{eq_tighten_constraint} and~\eqref{eq_tighten_constraint_2} are linear. 

\subsection{Projection of Solutions to Permutation Matrix}
After solving~\eqref{eq_max_traceWY_relaxed_tight} up to its global optimality using a convex solver, the remaining task is to project its solution back to the space of permutation matrices. This task can be done using different strategies~\cite{fogel2013convex}. In this work, we apply the linear assignment problem~\cite{burkard1999linear}, in which the projection can be computed efficiently by solving a linear program (LP). Specifically, let $\tilde{\bX}$ be the optimal solution of~\eqref{eq_max_traceWY_relaxed_tight}. The LP for projection can be formulated as 
\begin{equation}
\label{eq_lp_linear_assignment}
\begin{aligned}
& \max_{\bX} && \langle\bX, \tilde{\bX}\rangle, \\
&\text{subject to} &&  0 \le \bX_{i,j} \le 1 \;\;
 \sum_{i} \bX_{i,j} = 1, \;\;
 \sum_{j} \bX_{i,j} = 1,\\
\end{aligned}
\end{equation}
where $\langle\cdot,\cdot\rangle$ represents the inner product of two matrices.

The solution of~\eqref{eq_lp_linear_assignment} provides us with a set of $N$ correspondences. However, in~\eqref{eq_permuataion}, we only want to pick $m$ pairs. To tackle this with a simple heuristic, observe that in the optimal solution $\bX^*$ of~\eqref{eq_permuataion}, only $m$ rows/columns of $\bX^*$ contain the value of $1$, while the rest of the rows/columns contain all zeros. Therefore, from correspondence set obtained from~\eqref{eq_lp_linear_assignment}, we associate each pair of correspondence $(\bp_i,\bq_j)$ with a score that is equal to $\tilde{\bX}_{i,j}$. Finally, $m$ pairs with the highest scores are chosen as the approximate solution for~\eqref{eq_permuataion}. This approach has been empirically shown to be very efficient throughout our experiments.

\subsection{Main Algorithm}
\label{sec_sampling_strategy}
\begin{algorithm}[ht]\centering
\caption{SDRSAC}
\label{alg_sdrsac}                         
\begin{algorithmic}[1]                   
	\REQUIRE Input data $\cS$ and $\cD$, \ttmaxiter, \ttinneriter,  size of sampled subsets $N_{\text{sample}}$
	\STATE \ttit $\leftarrow 0$; \;\; \ttbestScore $\leftarrow$ 0;
	\WHILE{\ttit $<$ \ttmaxiter}
		\STATE $\cS' \leftarrow$  Randomly sample $N_{\text{sample}}$ points from ${\cS}$ \label{alg_line_pickS}
		\FOR{ $t=1$ to \ttinneriter}			
			\STATE $\cD' \leftarrow$  Randomly sample $N_{\text{sample}}$ points from ${\cD}$ \label{alg_line_pickD}
			\STATE \{$\cM, \bR, \bt$\} $\leftarrow$ \texttt{SDRMatching $(\cS, \cD, \cS',\cD')$} \; /*Described in Alg. \ref{alg_sdr_matching} */
			\IF{$|\cM| > $ \ttbestScore}
				\STATE \ttbestScore $\leftarrow |\cM|$; \;\;$\bR^* \leftarrow \bR$; \;\;$\bt^* \leftarrow \bt$
			\ENDIF
		\ENDFOR
		\STATE \ttit $\leftarrow$ \ttit + \ttinneriter
		\STATE  $T \leftarrow$ Number of iterations that satisfies the probabilistic stopping criteria (Sec.~\ref{sec_stopping})
		\IF {\ttit $\ge T$ } \STATE \texttt{return} \ENDIF
	\ENDWHILE
	\RETURN Best transformation ($\bR^*$, $\bt^*$), \ttbestScore 
\end{algorithmic}
\end{algorithm}

\begin{algorithm}[ht]\centering
	\caption{\texttt{SDRMatching}}
	\label{alg_sdr_matching}                         
	\begin{algorithmic}[1]                   
		\REQUIRE Input data $\cS$ and $\cD$, sampled subsets $\cS'$ and $\cD'$, threshold $\epsilon$.
		\STATE $\bA\leftarrow$ Matrix generated using Eq.~\eqref{eq_def_weight_matrix} with $\cP = \cS'$ and $\cQ = \cD'$ \label{step_generate_A}
		\STATE $\tilde{\bX} \leftarrow$ Solve~\eqref{eq_max_traceWY_relaxed_tight} with $\bA$ generated from Step~\ref{step_generate_A}. \label{step_solve_sdp}
		\STATE ${\bX} \leftarrow$ Solve~\eqref{eq_lp_linear_assignment} with $\tilde{\bX}$ obtained from Step~\ref{step_solve_sdp}.
		\STATE $\cM' \leftarrow \{ (\bs'_i \in \cS',\bd'_j \in \cD') | \bX_{i,j} = 1 \}$  
		\STATE $(\tilde{\bR}, \tilde{\bt}) \leftarrow $ Estimate transformation  based on correspondence set $\cM'$~\cite{eggert1997estimating,kabsch1976solution}.
		\STATE $(\bR, \bt) \leftarrow$ Refinement using ICP, initialized by $(\tilde{\bR}, \tilde{\bt})$
		\STATE $\cM \leftarrow \{(\bs_i \in \cS,\bd_j \in \cD) \;|\; \|\bR\bs_i + \bt - \bd_j\| \le \epsilon \}$\;\; /*$\bd_j$ is defined in~\eqref{eq_nearest_point} */
		\RETURN $\cM$, $\bR$, $\bt$
	\end{algorithmic}
\end{algorithm}

Although~\eqref{eq_max_traceWY_relaxed_tight} is convex, solving it for large value of $N$ is inefficient since the lifted variable $\bY$ belongs to the domain of $\bbR^{N^2\times N^2}$. However, for a small value of $N$ (typically, $N \le 20$), this problem can be optimized efficiently and~\eqref{eq_max_traceWY_relaxed_tight} provides a good approximation for the original problem~\eqref{eq_max_traceWY}. Taking advantage of this observation, one can develop a sampling approach where the number of points in each sample can be of any size, up to the limit that the employed solver can handle~\eqref{eq_max_traceWY_relaxed_tight} efficiently. More specifically, at each iteration, two subsets of points $\cS' \subseteq \cS$ and $\cD' \subseteq \cD$ are randomly sampled from $\cS$ and $\cD$, respectively. The cardinality of $\cS'$ and $\cD'$ can be controlled depending on the capability of the employed convex solver.  By solving~\eqref{eq_max_traceWY_relaxed_tight}, the outliers in the sampled subsets can be rejected and the best subset in each sample is retained for estimating the transformation. Like other randomized paradigms, this process can be repeated after a fixed number of iterations or until the stopping criteria is satisfied. The insight behind this strategy is that by sampling large subsets at each iteration, one is more likely to encounter the right subsets that contains  inliers, since if any sampled subset is contaminated, the outliers can be rejected efficiently by solving the SDP approximation discussed in Section~\ref{sec_SDP}. Algorithm~\ref{alg_sdrsac} summarizes our method.

\subsection{Stopping Criterion} 
\label{sec_stopping}
We follow~\cite{aiger20084} to derive our stopping criterion. Let us denote by $p_{I}$ be the probability of selecting one inlier (correct correspondence pair), and by $T$ the number of trials. Note that from statistics~\cite{forsyth2018probability}, the expected inlier rate in a sample of size $N_\text{sample}$ is also $p_{I}$. Moreover, after running Alg.~\ref{alg_sdr_matching}, only $m$ pairs of correspondences remain, allowing us to compute the stopping criterion based on $m$. Denote by $p_f$ the probability that the algorithm fails to find an all-inlier correspondence set after $T$ trials, $p_f$ can be computed as
\begin{equation}
	\label{eq_prob_fail}
	p_f = (1 - p_{I}^m)^T,
\end{equation} 
Therefore, in order to get a success probability to be greater than $p_s$, we must have the number of iterations greater than
\begin{equation}
	\label{eq_T}
	T \ge \frac{\log(1-p_s)}{\log(1-p_I^m)}.
\end{equation} 
Since the real inlier rate $p_I$ is not known in advance, following common practice of several randomized methods, we update this value during the sampling process, i.e., $p_I$ is iteratively updated using the inlier ratio of the current best-so-far solutions.

\section{Experimental Results}
To evaluate the performance of our proposed algorithm (SDRSAC), we conduct experiments on multiple sets of synthetic and real datasets and compare our approach against several state-of-the-art algorithms that can be used to solve point cloud registration without correspondences, including  ICP~\cite{besl1992method}, Trimmed ICP (TrICP)~\cite{chetverikov2002trimmed}, Iteratively Re-weighted Least Squares (IRLS)~\cite{bergstrom2014robust}, 4PCS~\cite{aiger20084} and its improvement Super4PCS~\cite{mellado2014super}. Within the class of globally optimal algorithms, we also compare SDRSAC against Go-ICP~\cite{yang2016go} and its robust version with trimming (TrGoICP). Note that conventional RANSAC method~\cite{fischler1981random} performs poorly for this type of problem, and in many cases it turns into a brute-force type algorithm. Thus, to save space, we only show results from established methods listed above. 

As we focus on validating the effectiveness of robust global registration, throughout the experiments, we measure and report the number of matches (objective function of~\eqref{eq_prob_def}) and run time for each method. 

All experiments are executed on a standard Ubuntu machine with $16$GB of RAM. SDRSAC, ICP, TrimmedICP were implemented using MATLAB. For 4PCS, Super4PCS~\cite{mellado2014super}, Go-ICP~\cite{yang2016go}, we use the released C++ code and the parameters suggested by the authors. All results of randomized methods are reported by averaging the outcomes obtained from $20$ different runs. In the following, we only report representative results. More results, implementation details and extensions to the case with known correspondences can be found in the supplementary material. 

\subsection{Synthetic Data}
\label{sec:synthetic_exp}
We first evaluate the performance of SDRSAC on synthetically generated data. The bunny point cloud from the Stanford dataset\footnote{http://graphics.stanford.edu/data/3Dscanrep/} is loaded and uniformly sampled to yield a source point cloud $\cS$ containing $N_s = 10,000$ points. To generate the target set $\cD$, we apply a random transformation $(\tilde{\bR}, \tilde{\bt})$ to $\cS$. Each point in $\cD$ is then perturbed with a Gaussian noise of zero mean and variance of $\sigma_{\text{noise}}=0.01$. To simulate partial overlapping, we randomly pick and remove $r\%$ of the points in $\cD$. In order to evaluate the performance of the algorithms with different outlier ratios, we repeat the experiments with $r=10, 15, \dots, 50\%$. The threshold $\epsilon$ in~\eqref{eq_rho} was chosen to be $0.01$ for all the methods. For SDRSAC, we choose the sample size to be \Nsample$=16$, and $m=4$. (The choice of \Nsample\ and $m$ will be studied in Sec.~\ref{sec_ablation_studies}).
\begin{figure*}[ht]
	\centering
	\includegraphics[width=0.45\textwidth]{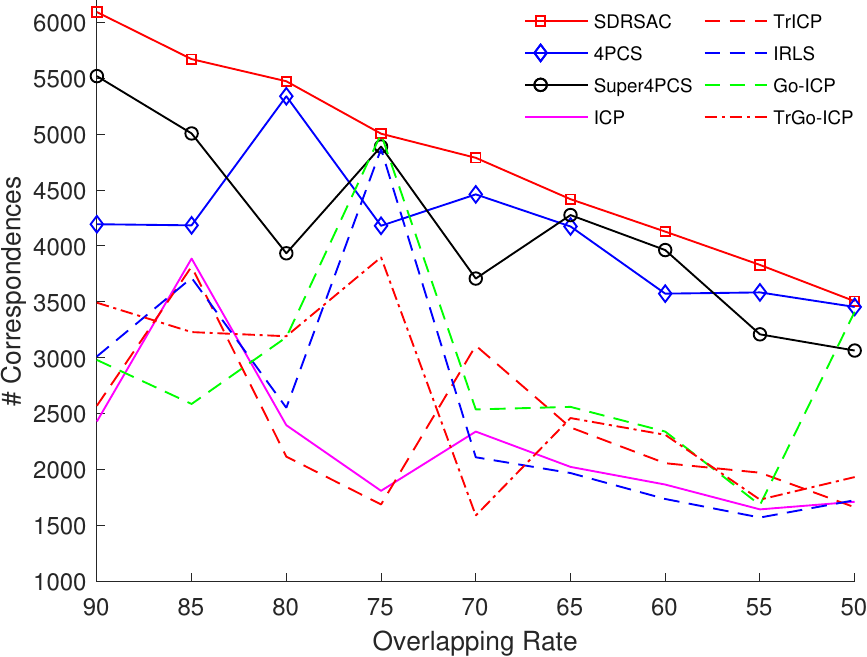} \hspace{0.1in}
	\includegraphics[width=0.45\textwidth]{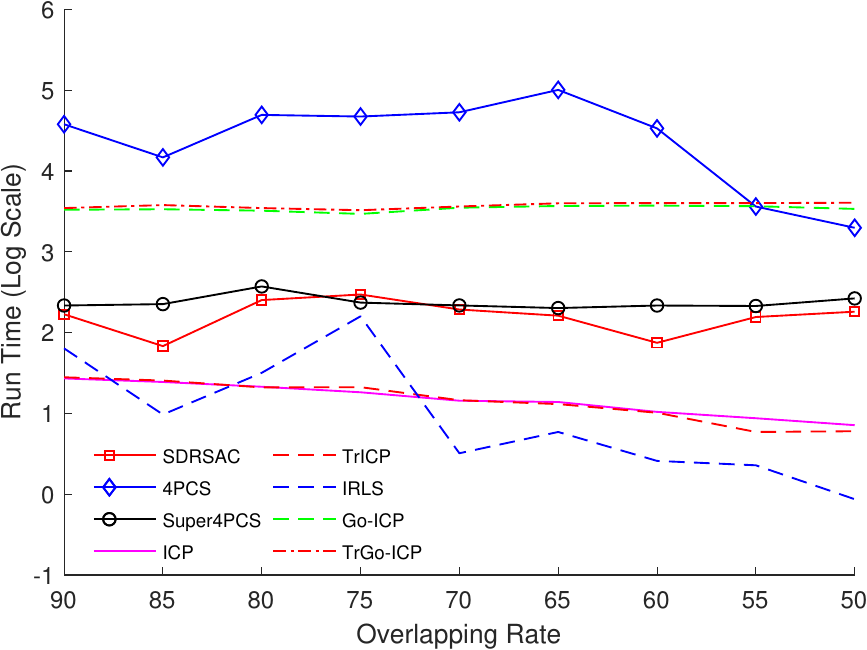}
	\caption{Results for experiments with synthetic data. Left: Number of correspondences (Objective of~\eqref{eq_prob_def}). Right: Run time (in log scale). }
	\label{fig_synthetic_results}
\end{figure*}
Figure~\ref{fig_synthetic_results} shows the number of matches (inliers) and run time for all the methods. It is evident that SDRSAC outperforms other methods in terms of correspondences obtained with faster (or comparable) execution time. Note that the run time of SDRSAC is very close to Super4PCS, though SDRSAC consistently attains higher solution quality. The performance of ICP, as anticipated, is unstable due to the effect of initialization (i.e., wrong initialization may lead to poor convergence), as shown in Fig.~\ref{fig_synthetic_results}. Moreover, as the outlier ratio increases, the consensus sizes produced by most ICP-based methods are poor due to their non-robustness.

\subsection{Real Data}
\begin{table*}
    \centering
    \resizebox{0.95\textwidth}{!}{        
        \begin{tabular}{|c |c |c |c |c |c |c |c |c |c |}
        \hline\hline        
        \multicolumn{2}{|c|}{Pairs} & SDRSAC & 4PCS & S-4PCS & \;ICP\; & TrICP & IRLS & GoICP & TrGoICP 
        \\ 
        \hline\hline
        
& \#Corrs  & \textbf{6990} & 5922 & 6388 & 2433 & 4455 & 2429 & 2664 & 6828\\ 
Office1\ 1,2 & Time(s) & 8.69 & 11.18 & 10.05 & 4.69 & 4.35 & 3.16 & 50.9 & 33.4\\ \hline 

& \#Corrs  & \textbf{4353} & 4184 & 2901 & 2132 & 1205 & 2125 & 2253 & 4101\\ 
Office1 8, 9 & Time(s) & 9.53 & 15.08 & 15.06 & 4.44 & 4.45 & 4.06 & 63.2 & 38.3\\ \hline

& \#Corrs  & \textbf{4992} & 4954 & 4933 & 655 & 487 & 673 & 1513 & 1413\\ 
Office1\ 15, 16 & Time(s) & 7.56 & 15.95 & 15.45 & 4.35 & 4.63 & 4.19 & 60.3 & 37.6\\ \hline 

& \#Corrs  & \textbf{4490} & 2998 & 3714 & 60 & 612 & 178 & 554 & 848\\ 
Office1\ 51, 52 & Time(s) & 10.05 & 15.45 & 15.03 & 4.29 & 4.55 & 4.39 & 60.5 & 38.5\\ \hline

& \#Corrs  & \textbf{5817} & 5602 & 4590 & 3609 & 4599 & 4016 & 4060 & 5655\\ 
Living1 1, 2 & Time(s) & 8.34 & 15.09 & 15.39 & 4.26 & 4.96 & 5.13 & 32.5 & 35.6\\ \hline

& \#Corrs  & \textbf{4768} & 3615 & 4327 & 1350 & 1364 & 4240 & 4398 & 4567\\ 
Living1\ 25, 26 & Time(s) & 9.27 & 15.07 & 15.95 & 4.34 & 4.42 & 4.35 & 65.3 & 36.1\\ \hline 

& \#Corrs  & \textbf{5558} & 5155 & 5433 & 4104 & 4783 & 4009 & 5120 & 5250\\ 
Living1\ 54, 55 & Time(s) & 9.2 & 15.05 & 15.65 & 4.29 & 4.32 & 5.19 & 61.2 & 35.9\\ \hline

& \#Corrs  & \textbf{5570} & 5459 & 3269 & 2028 & 1519 & 2808 & 2334 & 5319\\ 
Living1\ 32, 33 & Time(s) & 8.32 & 15.15 & 15.23 & 4.65 & 4.44 & 5.1 & 34.6 & 33.3\\ \hline 

        \hline\hline
        \end{tabular}
    }
    \caption{Results for real data experiments. For each pairs, the first row is the number of correspondences (\#Corrs) and the second row shows the run time in second. Note that S-4PCS represent Super4PCS. }
    \label{table_real_results}
\end{table*}
\begin{figure}
	\includegraphics[width=0.42\textwidth]{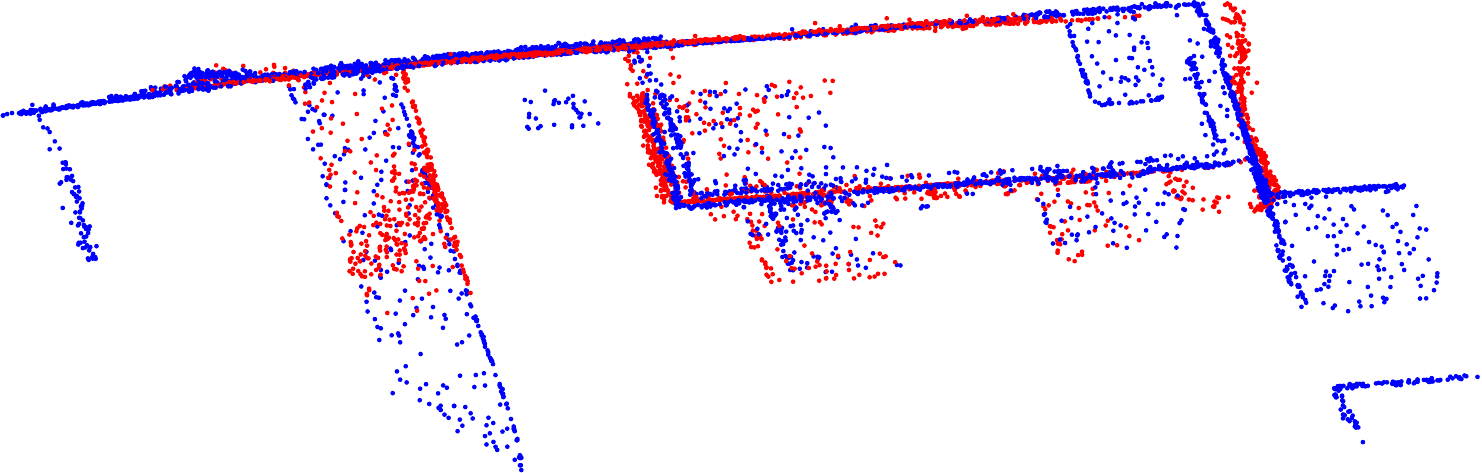} \hspace{0.1in}
	\includegraphics[width=0.42\textwidth]{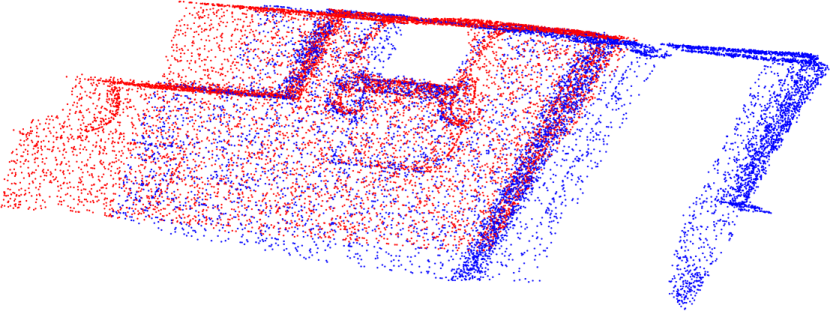}\\
	\caption{Examples of alignments using SDRSAC. From top to bottom: Office 8 and Office 9; LivingRoom 1 and LivingRoom 2. }
	\label{fig_alignment_example}
\end{figure}
In this section, we evaluate the performance of our proposed method on real datasets and compare it with existing approaches. The input point clouds for this experiment are obtained from the challenging Redwood 3D dataset~\cite{choi2015robust}. We randomly pick eight pairs of point clouds from the Office and Living Room repository, then uniformly sampled the point sets to obtain $10,000$ points on each input set. The input threshold $\epsilon$ in~\eqref{eq_rho} was chosen to be in the range of $0.01$ to $0.05$ and for each set of input data, $\epsilon$ was chosen to be the same for all benchmarking methods. 

Table~\ref{table_real_results} shows the number of matches and run time for each method. The sample qualitative results are  displayed in Figure~\ref{fig_alignment_example}. As can be seen, on average, SDRSAC obtains higher matches compared to other competitors. We also observe  that local methods such as ICP or IRLS can sometimes converge fast to very good results, but are unstable due to the effect of bad initializations. The robust version TrGo-ICP performs slightly better than Go-ICP, but also fails in some cases due to high outlier ratios residing in the input data. 

\subsection{Ablation Studies}
\label{sec_ablation_studies}
This section analyzes the effect of different parameter settings to the performance of our proposed method, and suggest the choices for the hyperparameters.

\subsubsection{Effect of Sample Size $N_\text{sample}$}
\begin{figure}[ht]
    \centering
    \includegraphics[width = 0.32\columnwidth]{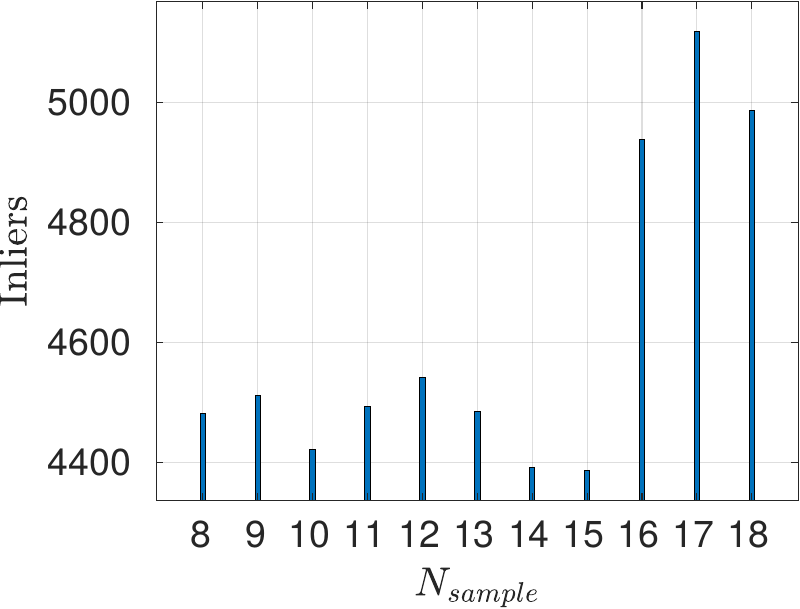}
    \includegraphics[width = 0.32\columnwidth]{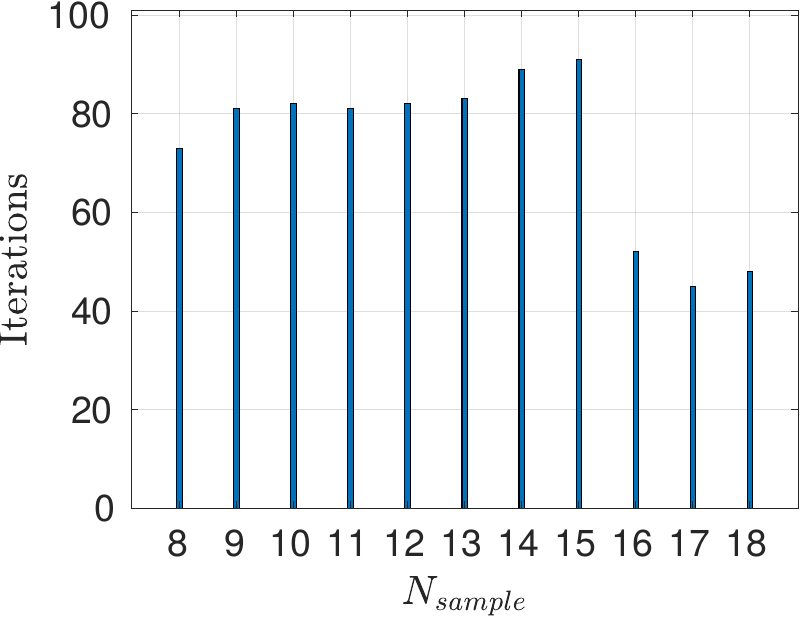}
    \includegraphics[width = 0.32\columnwidth]{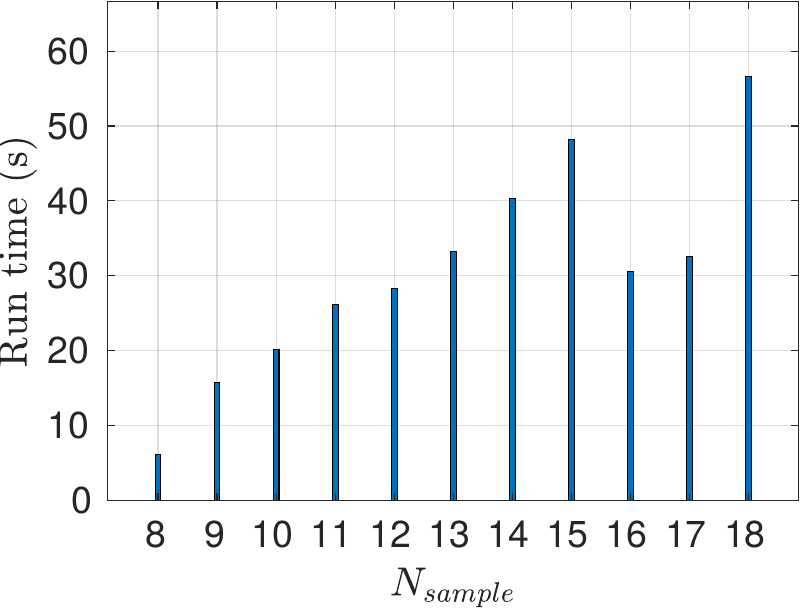}
    \caption{Performance analysis for different values of $N_\text{sample}$. Left: Consensus size, Middle:\# Iterations, Right: Run time.}
    \label{fig:study_N}
\end{figure}
Fig.~\ref{fig:study_N} plots the accuracy (consensus size), number of iterations (to satisfy the stopping criterion (14)), and the total run time when \Nsample~increases from $8$ to $18$ and $m=4$ for a synthetic dataset (generated as described in Sec.~\ref{sec:synthetic_exp}) with $N=10,000$ and $p_I \approx 30\%$ (the plotted results are the median over $20$ runs). Our obtained results conform to statistical theory~\cite{forsyth2018probability}.
Specifically, the inlier rates in the subsets of \Nsample~points have a mean of $p_I$ and standard deviation (SD) of $\sqrt{\frac{p_I(1-p_I)}{N_\text{sample}}}$. From Fig.~\ref{fig:study_N}, we observe that for small \Nsample\ (\Nsample $< 16$), the obtained consensus sizes are low and the algorithm requires a large number of iterations. In particular, if \Nsample~ is small, then SD of the inlier rates is large among sampled subsets, thus some subsets may be contaminated with large numbers of outliers, which affects the quality and stability of our algorithm. When \Nsample~reaches around $16$, SD is $\sqrt{\frac{0.3*0.7}{16}}\approx0.1$, hence the  inlier rates vary only slightly among subsets, leading to a stable performance, as shown in Fig.~\ref{fig:study_N} (left, middle). While it is good to use large \Nsample~ from the statistical point of view, SDP solver would take  more time at each iteration for large \Nsample~, resulting in  longer overall run time as shown in Fig.~\ref{fig:study_N} (right). 
Empirically, we found that $N_\text{sample}=16$ provides a good trade-off between algorithm stability (accuracy) and run time in many settings.


\subsubsection{Effect of $m$}
\begin{figure}[ht]
	\centering
	\includegraphics[width = 0.4\columnwidth]{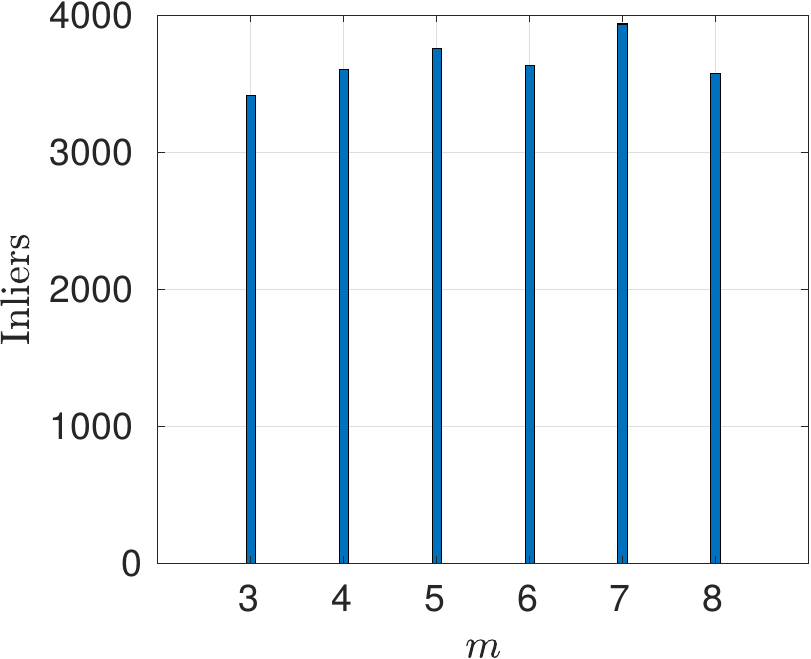}
	\includegraphics[width = 0.4\columnwidth]{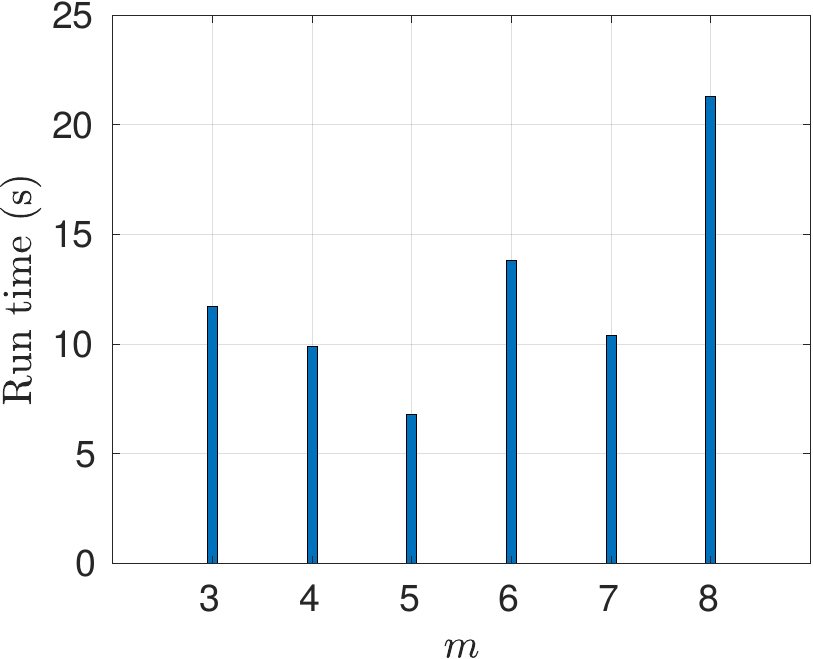}
	\caption{Plots of consensus sizes (left) and  run times (right)  over different values of $m$.}
	\label{fig_runtime_m}
\end{figure}

This experiment is conducted to study on the effect of $m$ (introduced in Sec.~\ref{sec_SDP}) on the solution quality and run time of our approach. We repeat the experiment for a synthetic dataset containing $N=5,000$ points per point cloud with $10\%$ outliers. The sample size $N_\text{sample}$ is set to $16$. All results are obtained as median over $20$ runs. Fig.~\ref{fig_runtime_m} (left) plots the number of inliers obtained at termination (using the stopping criterion~\eqref{eq_T}). Observe that the performance is quite stable at different values of $m$, which demonstrates the effectiveness of SDP to reject outliers and the strengths of our proposed sampling scheme.
In Fig.~\ref{fig_runtime_m} (right), we also plot the run time required until termination. Apparently, although the solution qualities for different values of $m$ are similar as discussed above, the run time is affected by $m$. Specifically, as $m$ increases, SDRSAC takes more iterations before the stopping criterion~\eqref{eq_T} is met. This can be explained by recalling from~\eqref{eq_prob_fail} that large $m$ increase the failure probability $p_f$. Moreover, for any sampled subset, the expected number of inliers is ${N_\text{sample}} \times p_I$, hence if $m$ is larger than this value, outliers may be included. In our experiments, the choice of $m=4$ works best in most scenarios (the minimal case of $m=3$ may not be very stable due to noise in 3D data points and the 3 correspondences could be near each other spatially).	

\section{Conclusions}
We have presented a novel and efficient randomized approach for robust point cloud registration without correspondences. Our method is based on a novel utilization of graph matching formulation for the correspondence problem, together with a novel cost function and a tight SDP relaxation scheme. We embed the formulation into a new sampling strategy which samples larger-than-minimal subsets. Extensive experiments with several synthetic and real datasets show that our algorithm is competitive to existing state-of-the-art approaches. Our framework can also be extended for the problems with known correspondences.

\section*{Acknowledgement}
This work was supported in part by both ST Electronics and the National Research Foundation(NRF), Prime Minister's Office, Singapore under Corporate Laboratory @ University Scheme (Programme Title: STEE Infosec - SUTD Corporate Laboratory).

{\small
\bibliographystyle{ieee}
\bibliography{main}
}

\vfill
\pagebreak
\appendix
\section*{Supplementary Material}
\section{Extension of the algorithm to the case of known correspondences}

As mentioned in the main paper, our algorithm (SDRSAC) can be extended easily to registration problems with known correspondences. Such extension can be done easily with a slight modification to the original algorithm. In this section, we discuss in details a new algorithm to enable SDRSAC for problems where putative correspondences are known (we call the new algorithm CSDRSAC -- SDRSAC with correspondences). Also, we will provide some preliminary experiment results where show that CSDRSAC performs much better than RANSAC~\cite{fischler1981random}.

\subsection{Algorithm}
The main idea of the extension is to make use of the information provided by the a priori putative set of correspondences to obtain the subset $\cD'$, instead of sampling from $\cD$ (Line 5 in the SDRSAC algorithm described in the main paper). Specifically, the algorithm can be described as Alg.~\ref{alg_sdrsac}

\addtocounter{algorithm}{2}
\begin{algorithm}[!htbp]
	\begin{algorithmic}[1]                   
		\REQUIRE Input data $\cS$ and $\cD$, \ttmaxiter,  size of sampled subsets $N_{\text{sample}}$
		\STATE \ttit $\leftarrow 0$; \;\; \ttbestScore $\leftarrow$ 0;
		\WHILE{\ttit $<$ \ttmaxiter}
		\STATE $\cS' \leftarrow$ Randomly sample from $\cS$ with $|\cS'| = N_{\text{sample}}$
		\STATE $\cD' \leftarrow$ Correspondences of $\cS'$ where $\cD'\subseteq \cD$
		
		\STATE \{$\cM, \bR, \bt$\} $\leftarrow$ \texttt{SDRMatching $(\cS, \cD, \cS',\cD')$} \; /*As Alg.2 in main paper */
		\IF{$|\cM| > $ \ttbestScore}
		\STATE \ttbestScore $\leftarrow |\cM|$; \;\;$\bR^* \leftarrow \bR$; \;\;$\bt^* \leftarrow \bt$
		\ENDIF
		
		\STATE \ttit $\leftarrow$ \ttit + 1
		\STATE  $T \leftarrow$ Number of iterations that satisfies the stopping criterion.
		\IF {\ttit $\ge T$ } \STATE \texttt{return} \ENDIF
		\ENDWHILE
		\RETURN Best transformation($\bR^*$, $\bt^*$), \ttbestScore 
	\end{algorithmic}
	\caption{CSDRSAC}
	\label{alg_sdrsac} 
\end{algorithm}

\subsection{Experiments}
\begin{figure}[ht]
	\centering
	\includegraphics[width=0.3\textwidth]{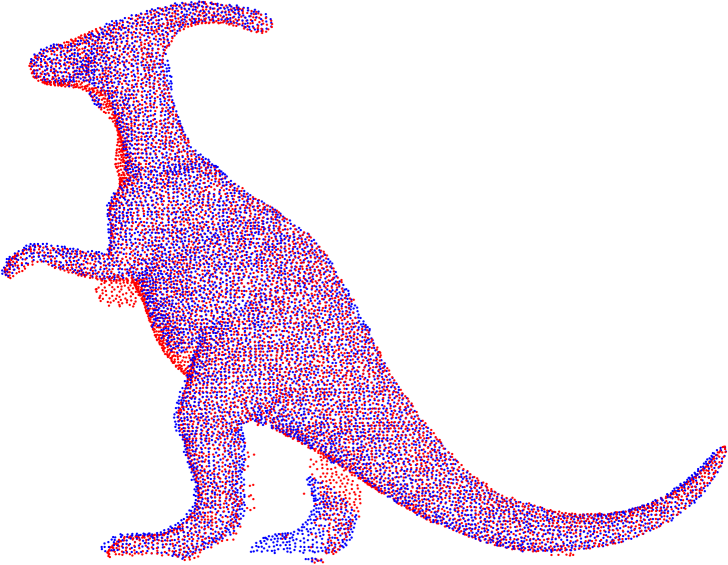}
	\includegraphics[width=0.3\textwidth]{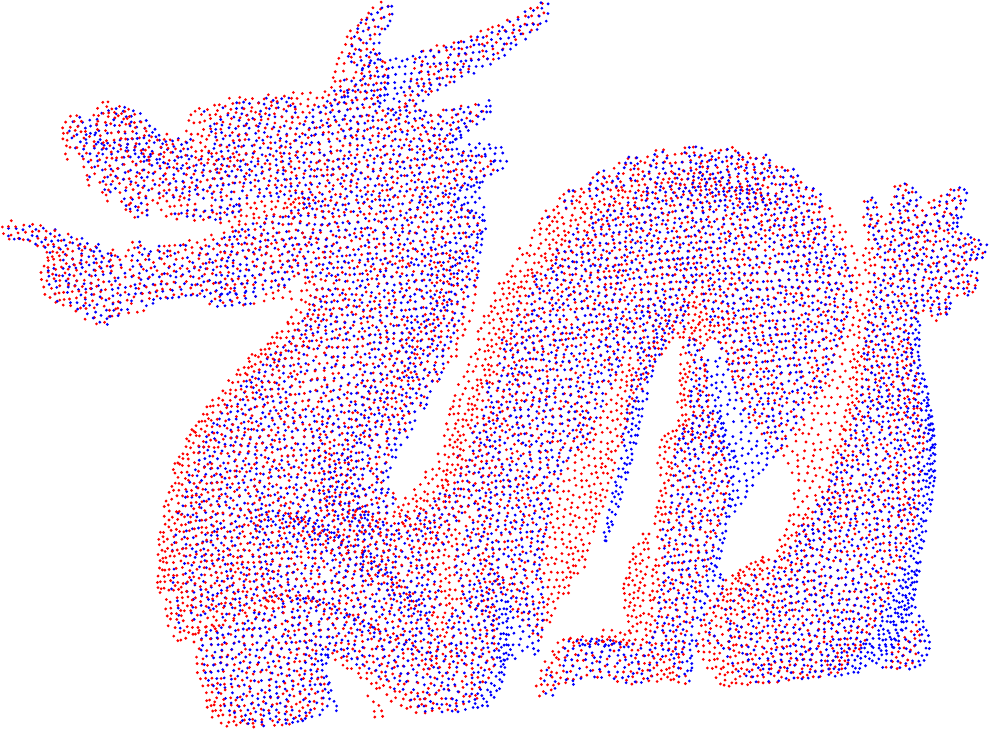}
	\includegraphics[width=0.3\textwidth]{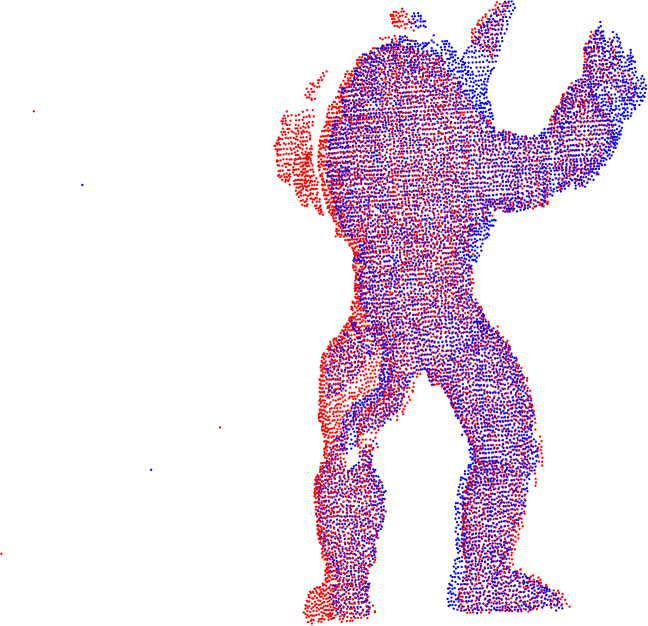}
	\caption{Examples of point clouds aligned by CSDRSAC. From top to bottom: T-rex; Dragon; Armadillo}
	\label{fig_csdrsac}
\end{figure}

\begin{table*}[ht]
	\centering
	\begin{tabular}{|c|c|c|c|c|c|c|c|}
		\hline
		& & Bunny & Armadillo & Dragon & Buddha & Chicken & T-rex \\ \hline
		\multirow{2}{*}{SDRSAC} & \#Corrs & \textbf{6850}    &  \textbf{6898}  &  \textbf{6828}  & \textbf{6739}   &  \textbf{7260} & \textbf{6531} \\ \cline{2-8} 
		& Time (s)   & 14.56   & 15.73   &  15.28  & 13.46   &   16.65 & 12.25 \\ \hline
		\multirow{2}{*}{RANSAC} & \#Corrs & 6530   &  6793  &  6818   &  6695  & 6956  & 6521 \\ \cline{2-8} 
		& Time (s)   & 195.5    &  420.27  &   153.93 &  445.37  & 156.61  & 352.52 \\ \hline
	\end{tabular}
	\caption{Experiment results for CSDRSAC and RANSAC. For each pair of input data, $N=2000$ key points were used for registration}
	\label{table_csdr_results}
\end{table*}
In this section, we compare CSDRSAC against RANSAC~\cite{fischler1981random}. For input data, we use the Standford 3D dataset and the UWA datasets. The keypoints were generated and matched using the data and code provided by~\cite{bustos2015guaranteed}. For each pair of shapes, a set of $N = 2000$ putative correspondences are supplied to the algorithms.   The number correspondences and run time for five pairs are shown in the Table~\ref{table_csdr_results} and the alignment results are displayed in Fig.~\ref{fig_csdrsac}. Note that the number of correspondences are measured based on the original point clouds instead of the feature set. Apparently, CSDRSAC performs much better than RANSAC. This suggest that CSDRSAC is a promising method, which deserves further investigation to develop better randomized algorithm for registration problems with known correspondences.

\section {More experiments on registration problems without correspondences}
\begin{table*}[ht]
\centering
\begin{tabular}{|c|c|c|c|c|c|c|c|c|c|}
\hline
\multicolumn{2}{|c|}{Pairs} & SDRSAC & 4PCS & S-4PCS & \;ICP\; & TrICP & IRLS & GoICP & TrGoICP 
\\ 
\hline\hline
Office2\_1 & \#Corrs  & \textbf{8962} & 7644 & 8335 & 8505 & 8615 & 8575 & 953 & 5685\\ \cline{2-10}
Office2\_2 & Time(s) & 10.15 & 10.52 & 10.68 & 4.32 & 5.15 & 11.15 & 40.4 & 35.5\\ \hline
Office2\_5 & \#Corrs  & \textbf{5630} & 4337 & 4301 & 1887 & 3206 & 4976 & 3813 & 2811\\ \cline{2-10}
Office2\_6 & Time(s) & 8.65 & 10.52 & 10.35 & 4.26 & 4.65 & 12.53 & 30.1 & 28.5\\ \hline
Office2\_10 & \#Corrs  & \textbf{5975} & 5604 & 5275 & 1881 & 2714 & 2272 & 2840 & 3338\\ \cline{2-10}
Office2\_11 & Time(s) & 7.39 & 10.19 & 10.35 & 4.48 & 5.25 & 22.5 & 29.5 & 28.5\\ \hline
Living2\_20 & \#Corrs  & \textbf{3787} & 3662 & 3347 & 2227 & 2368 & 2267 & 1990 & 3300\\ \cline{2-10}
Living2\_21 & Time(s) & 8.65 & 10.25 & 10.12 & 4.65 & 4.13 & 4.45 & 32.5 & 29.3\\ \hline
Living2\_5 & \#Corrs  & \textbf{3862} & 3523 & 3545 & 1358 & 1456 & 1286 & 1618 & 2553\\ \cline{2-10}
Living2\_6 & Time(s) & 6.8 & 10.25 & 10.65 & 4.35 & 4.56 & 8.78 & 33.5 & 30.2\\ \hline
Living2\_47 & \#Corrs  & \textbf{2892} & 2788 & 2392 & 587 & 614 & 374 & 379 & 1688\\ \cline{2-10}
Living2\_48 & Time(s) & 11.85 & 15.95 & 15.23 & 4.56 & 4.67 & 29.75 & 39.5 & 42.6\\ \hline

\end{tabular}
\caption{Results for real data experiments. For each pairs, the first row is the number of correspondences (\#Corrs) and the second row shows the run time in second. Note that S-4PCS represent Super4PCS}
\label{table_real_results}
\end{table*}

In this section, we provide more results for registration problem without correspondences. These experiments were setup with the same settings as described in Section 4 in the main paper. The results are shown in Table~\ref{table_real_results}. As can be seen in Table~\ref{table_real_results}, our method consistently provides comparable results compared to other state-of-the-art methods on point cloud registration without correspondences.

\vfill
\end{document}